\documentclass[10pt,twocolumn,letterpaper]{article}

\usepackage[pagenumbers]{cvpr} 

\usepackage[dvipsnames]{xcolor}

\usepackage{multirow, mathtools, cuted} 

\definecolor{cvprblue}{rgb}{0.21,0.49,0.74}
\usepackage[pagebackref,breaklinks,colorlinks,citecolor=cvprblue]{hyperref}

\def\paperID{*****} 
\def\confName{CVPR}
\def\confYear{2024}

\title{VolumeDiffusion: Flexible Text-to-3D Generation with \\ Efficient Volumetric Encoder}

\author{
    Zhicong Tang$^{1}$
    \quad
    Shuyang Gu$^{2}$
    \quad
    Chunyu Wang$^{2}$
    \\
    Ting Zhang$^{2}$
    \quad
    Jianmin Bao$^{2}$
    \quad
    Dong Chen$^{2}$
    \quad
    Baining Guo$^{2}$
    \\
    $^1$Tsinghua University
    \quad
    $^2$Microsoft Research
    \\
    {\tt\small tzc21@mails.tsinghua.edu.cn}
    \\
    {\tt\small \{shuyanggu,chnuwa,ting.zhang,jianbao,doch,bainguo\}@microsoft.com}
}

\begin{document}
\maketitle

\begin{abstract}
This paper introduces a pioneering 3D volumetric encoder designed for text-to-3D generation.
To scale up the training data for the diffusion model, a lightweight network is developed to efficiently acquire feature volumes from multi-view images. The 3D volumes are then trained on a diffusion model for text-to-3D generation using a 3D U-Net. 
This research further addresses the challenges of inaccurate object captions and high-dimensional feature volumes. The proposed model, trained on the public Objaverse dataset, demonstrates promising outcomes in producing diverse and recognizable samples from text prompts. Notably, it empowers finer control over object part characteristics through textual cues, fostering model creativity by seamlessly combining multiple concepts within a single object. This research significantly contributes to the progress of 3D generation by introducing an efficient, flexible, and scalable representation methodology.
\end{abstract}
\vspace{-0.21cm}
\section{Introduction}
\label{sec:intro}

Text-to-image diffusion models~\cite{rombach2022high} have seen significant improvements thanks to the availability of large-scale text-image datasets such as Laion-5B~\cite{schuhmann2022laion}. This success suggests that scaling up the training data is critical for achieving a ``stable diffusion moment'' in the challenging text-to-3D generation task.  To achieve the goal, we need to develop a 3D representation that is \emph{efficient} to compute from the massive data sources such as images and point clouds, and meanwhile \emph{flexible} to interact with text prompts at fine-grained levels.

Despite the increasing efforts in 3D generation, the optimal representation for 3D objects remains largely unexplored. Commonly adopted approaches include Tri-plane~\cite{wang2023rodin,gupta20233dgen} and implicit neural representations (INRs)~\cite{jun2023shap}. However, Tri-plane have been only validated on objects with limited variations such as human faces due to the inherent ambiguity caused by factorization. The global representation in INR makes it hard to interact with text prompts at the fine-grained object part level, constituting a significant limitation for generative models.

In this work, we present a novel 3D volumetric representation that characterizes both the texture and geometry of small parts of an object using features in each voxel, similar to the concept of pixels in images. Differing from previous approaches such as~\cite{long2022sparseneus,chen2021mvsnerf}, which require additional images as input, our method allows us to directly render images of target objects using only their feature volumes.
Meanwhile, the feature volumes encode generalizable priors from image features, enabling us to use a shared decoder for all objects. The above advantages make the representation well-suited for generation tasks. 

To scale up the training data for the subsequent diffusion model, we propose a lightweight network to efficiently acquire feature volumes from multi-view images, bypassing the expensive per-object optimization process required in previous approaches~\cite{wang2023rodin}. In our current implementation, this network can process 30 objects per second on a single GPU, allowing us to acquire $500K$ models within hours. It also allows extracting ground-truth volumes on-the-fly for training diffusion models which eliminates the storage overhead associated with feature volumes. In addition to the efficiency, this localized representation also allows for flexible interaction with text prompts at fine-grained object part level. This enhanced controllability paves the way for creative designs by combining a number of concepts in one object.

We train a diffusion model on the acquired 3D volumes for text-to-3D generation using a 3D U-Net~\cite{ronneberger2015u}. This is a non-trivial task that requires careful design. First, the object captions in the existing datasets~\cite{deitke2023objaverse,deitke2023objaversexl} are usually inaccurate which may lead to unstable training if not handled properly. To mitigate their adverse effects, we carefully designed a novel schedule to filter out the noisy captions, which notably improves the results. Second, the feature volumes are usually very high-dimensional, \eg C $\times$ $32^3$ in our experiments which potentially pose
challenges when training the diffusion model. We adopted a new noise schedule that shifted towards larger noise due to increased voxel redundancy. 
Meanwhile, we proposed the low-frequency noise strategy to effectively corrupt low-frequent information when training the diffusion model. We highlight that this structural noise has even more important effects than that in images due to the higher volume dimension. 

We train our model on the public dataset Objaverse~\cite{deitke2023objaverse} which has $800K$ objects ($100K$ after filtering). Our model successfully produces diverse and
recognizable samples from text prompts. Compared to Shap$\cdot$E~\cite{jun2023shap}, our model obtains superior results in terms of controlling the characteristics of object parts through text prompts, although we only use less than $10\%$ of the training data (Shap$\cdot$E trained on several million private data according to their paper). For instance, given the text prompt ``\textit{a black chair with red legs}'', we observe that Shap$\cdot$E usually fails to generate red legs. We think it is mainly caused by the global implicit neural representation which cannot interact with text prompts at fine-grained object part level. Instead, our localized volumetric representation, similar to images, can be flexibly controlled by text prompts at voxel level. We believe this is critical to enhance the model's creativity by combining a number of concepts in one object.

\section{Related Work}
\label{sec:related_work}

\subsection{Differentiable Scene Representation}

Differentiable scene representation is a class of algorithms that encodes a scene and can be rendered into images while maintaining differentiability. It allows scene reconstruction by optimizing multi-view images and object generation by modeling the representation distribution. It can be divided into implicit neural representation (INR), explicit representation (ER), and hybrid representation (HR).

Neural Radiance Field (NeRF)~\cite{mildenhall2021nerf} is a typical INR that encodes a scene as a function mapping from coordinates and view directions to densities and RGB colors. Densities and RGB colors of points along camera rays are integrated to render an image, and the function mapping can be trained to match the ground-truth views. While NeRF uses Multi-Layer Perceptron (MLP) to encode the underlying scene, its success gave rise to many follow-up works that explored different representations.

Plenoxels~\cite{fridovich2022plenoxels} is an ER that avoids a neural network decoder and directly encodes a scene as densities and spherical harmonic coefficients at each grid voxel. TensoRF~\cite{chen2022tensorf} further decomposes the voxel grid into a set of vectors and matrices as an ER, and values on each voxel are computed via vector-matrix outer products. 

Instant-NGP~\cite{muller2022instant} is an HR that uses a multi-resolution hash table of trainable feature vectors as the input embedding of the neural network decoder and obtains results with fine details. \cite{chan2022efficient} proposed a Tri-plane HR that decomposes space into three orthogonal planar feature maps, and features of points projected to each plane are added together to represent a point in space. DMTet~\cite{shen2021deep} is also an HR that combines a deformable tetrahedral grid and Signed Distance Function (SDF) to obtain a precise shape, and the underlying mesh can be easily exported by Marching Tetrahedra algorithm~\cite{doi1991efficient}.

\subsection{3D Generation}

Some recent works escalate to text-to-3D generation via the implicit supervision of pretrained text-to-image or vision-language models. \cite{jain2022zero} use a contrastive loss of CLIP~\cite{radford2021learning} text feature and rendered image feature to optimize a 3D representation. \cite{poole2022dreamfusion,wang2023prolificdreamer} develop the Score Distillation Sampling (SDS) method, leveraging the semantic understanding and high-quality generation capabilities of text-to-image diffusion models. \cite{tang2023make,yu2023hifi} further combine CLIP, SDS, reference view, and other techniques to push the quality.

Though these optimization-based methods yield outstanding visual fidelity and text-3D alignment, they suffer from the cost of time-consuming gradient back-propagation and optimization, which could take hours for each text prompt. Also, the Janus problem, \ie multiple faces on one object, the over-saturated color, the instability and sensitivity to random seed, and the lack of diversity arise from the distillation of text-to-image diffusion models.

Other works resort to directly generate 3D representations and apply explicit supervision. \cite{chan2022efficient,gao2022get3d,wei2023taps3d,huang2023textfield3d} train Generative Adversarial Networks (GANs)~\cite{goodfellow2014generative} on multi-view rendered images, which may be limited by the capability of generator and discriminator. \cite{cao2023large,muller2023diffrf,wang2023rodin} train text-conditioned diffusion models on pre-optimized and saved NeRF parameters in the cost of a time-consuming fitting preparation and expensive storage. \cite{lasdiffusion,nam20223d,qi2023vpp,yu2023pushing} fall back to a 2-stage manner of first generating geometry-only point clouds and then utilizing off-the-shelf texture painting methods. \cite{liu2023one,liu2023syncdreamer} rely on the 3D structural understanding of pretrained pose-conditioned diffusion model~\cite{liu2023zero}, and perform 3D reconstruction with generated multi-view images, which may be not 3D consistent. \cite{szymanowicz2023viewset,karnewar2023holodiffusion} study a simplified category-specific generation and are unconditional or conditioned on image input. Although \cite{jun2023shap} successfully maps text to 3D shapes at scale, they choose parameters of NeRF MLP as modeling representation, which may be highly non-linear and inflexible for fine-grained text prompts control.

\section{Method}
\label{sec:method}

\begin{figure*}[t]
  \centering
   \includegraphics[width=0.99\linewidth]{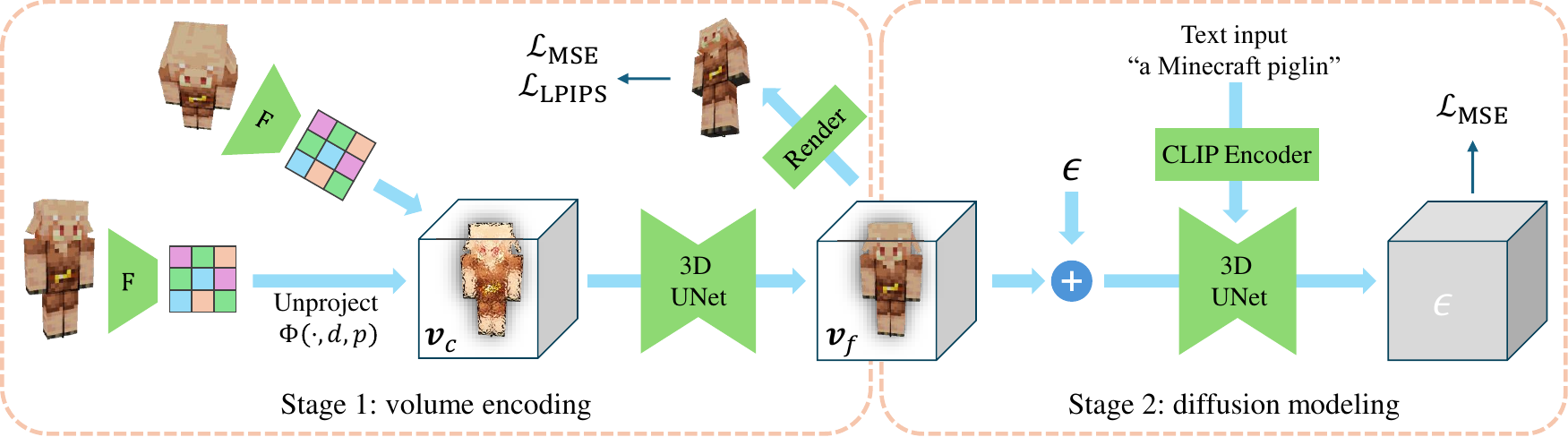}
   \caption{Framework of VolumeDiffusion. It comprises the volume encoding stage and the diffusion modeling stage. The encoder unprojects multi-view images into a feature volume and do refinements. The diffusion model learns to predict ground-truths given noised volumes and text conditions.}
   \label{fig:method}
\end{figure*}

Our text-to-3D generation framework comprises two main stages: the encoding of volumes and the diffusion modeling phase.
In the volume encoding stage, as discussed in Section~\ref{sec:volume_encoder}, we have chosen to use feature volume as our 3D representation and utilize a lightweight network to convert multi-view images into 3D volumes. The proposed method is very efficient and bypasses the typically costly optimization process required by previous methods, allowing us to process a substantial number of objects in a relatively short period of time.
In the diffusion modeling phase, detailed in Section~\ref{sec:diffusion_model}, we model the distribution of the previously obtained feature volumes with a text-driven diffusion model. This stage of the process is not without its challenges, particularly in relation to the high dimensionality of the feature volumes and the inaccuracy of object captions in the datasets. We have therefore developed several key designs to mitigate these challenges during the training process.

\subsection{Volume Encoder}
\label{sec:volume_encoder}

\subsubsection{Volume Representation}

One of the key points to train 3D generation models is the selection of appropriate 3D representations to serve as the latent space. The 3D representation should be able to capture the geometry and texture details of the input object and be flexible for fine-grained text control. Furthermore, the 3D representation should be highly efficient in obtaining and reconstructing objects for scalability.

Previous representations such as NeRF~\cite{mildenhall2021nerf}, Plenoxel~\cite{fridovich2022plenoxels}, DMTet~\cite{shen2021deep},  TensoRF~\cite{chen2022tensorf}, Instant-NGP~\cite{muller2022instant}, and Tri-plane~\cite{chan2022efficient} all have their limitations to serve as the latent space. For instance, the globally shared MLP parameters across different coordinates in NeRFs cause them inflexible and uncontrollable to local changes. Representations storing an explicit 3D grid, like Plenoxel and DMTet, require high spatial resolutions, resulting in large memory costs for detailed scene representation. TensoRF, Instant-NGP, and Tri-plane decompose the 3D grid into multiple sub-spaces with lower dimension or resolution to reduce memory costs but also introduce entanglements.

In this work, we propose a novel representation that merges a lightweight decoder with a feature volume to depict a scene. The lightweight decoder comprises a few layers of MLP, enabling high-resolution, fast, and low-memory cost rendering. The feature volume, instead of storing explicit values, houses implicit features and effectively reduces memory costs. The features of a spatial point are tri-linearly interpolated by the nearest voxels on the volume. The decoder inputs the interpolated feature and outputs the density and RGB color of the point. The feature volume is isometric to the 3D space, providing extensive controllability over each part of an object.

\subsubsection{Feed-forward Encoder}

Unlike previous works~\cite{cao2023large,muller2023diffrf,wang2023rodin} that iteratively optimize the representation for each object in a time-consuming way, we use an encoder that directly obtains the feature volume of any object within a forward pass.

As shown in Figure~\ref{fig:method}, the encoder takes a set of multi-view photos of an object $(\textbf{x}, \textbf{d}, \textbf{p})$, where $\textbf{x}, \textbf{d}\in\mathbb{R}^{N\times 3 \times H \times W}$ represents the image and depth of $N$ views, $\textbf{p} = \{p^{(i)}\}_{i=1}^{N}$ represents the corresponding camera parameters, including the camera poses and field of view (FOV). 
We first extract features from 2D images with a small network $\mathbf{F}$ composed of two layers of convolution.
We then unproject the features into a coarse volume $\textbf{v}_c$ according to depths and camera poses, \ie,
\begin{equation}
   \mathbf{v}_c=\Phi(\mathbf{F}(\textbf{x}), \textbf{d}, \textbf{p}),  
\end{equation}
where $\Phi$ represents the unproject operation. For each point on camera rays, we first calculate its distance to the camera, then obtain a weight $\mathbf{w}_i=\exp{(-\lambda\Delta d_i)}$ where $\Delta d_i$ is the difference of calculated distance and ground-truth depth. The feature of each voxel is the weighted average of features unprojected from different views.

Secondly, we apply a 3D U-Net~\cite{ronneberger2015u} module to refine the aggregated feature volume to produce a smoother volume
\begin{equation}
    \mathbf{v}_f = \Psi(\mathbf{v}_c).
\end{equation}
Then ray marching and neural rendering are performed to render images from target views. In the training stage, we optimize the feature extracting network, the 3D U-Net, and the MLP decoder end-to-end with $L_2$ and LPIPS~\cite{zhang2018unreasonable} loss on multi-view rendered images.

The proposed volume encoder is highly efficient for two primary reasons. Firstly, it is capable of generating a high-quality 3D volume with 32 or fewer images once it is trained. This is a significant improvement over previous methods~\cite{wang2023rodin}, which require more than 200 views for object reconstruction. Secondly, our volume encoder can encode an object in approximately 30 milliseconds using a single GPU. This speed enables us to generate $500K$ models within a matter of hours. As a result, there's no need to store these feature volumes. We extract ground-truth volumes for training diffusion models on-the-fly. It effectively eliminates the expensive storage overhead associated with feature volumes.

\subsection{Diffusion Model}
\label{sec:diffusion_model}

\begin{figure}[t]
  \centering
  \includegraphics[width=0.99\linewidth]{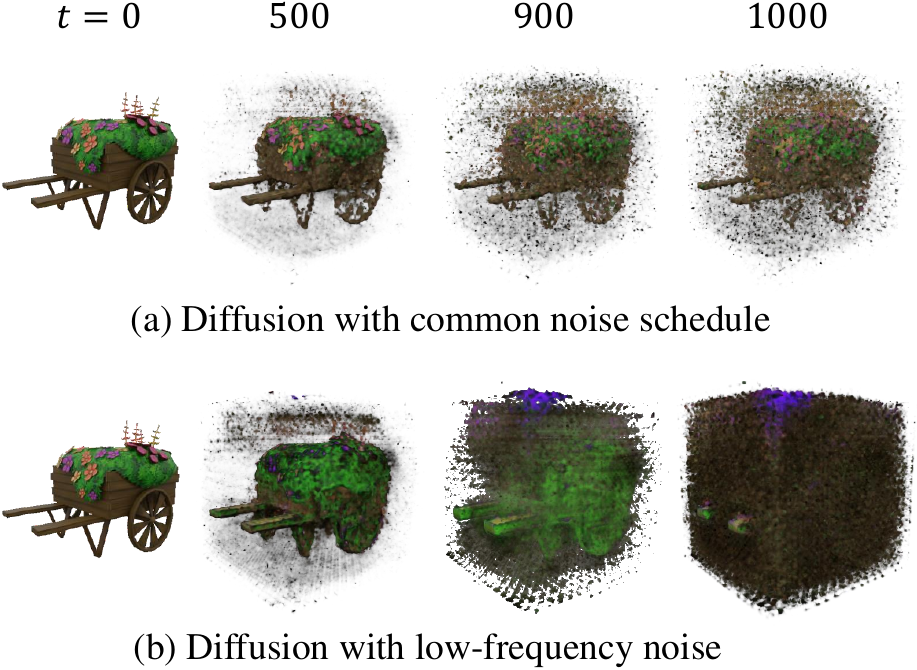}
  \caption{Renderings of noised volumes. Volumes with common \iid noise are still recognizable at large timesteps, while low-frequency noise effectively removes information.}
  \label{fig:noised_volume}
\end{figure}

\subsubsection{Devil in High-dimensional Space}
\label{sec:high_dimensional_space}

Unlike the conventional text-to-image diffusion models, our text-to-3D diffusion model is designed to learn a latent distribution that is significantly more high-dimensional. This is exemplified in our experiments where we utilize dimensions such as $C \times 32^3$, in stark contrast to the $4\times64^2$ employed in Stable Diffusion. This heightened dimensionality makes the training of diffusion models more challenging.

Figure~\ref{fig:noised_volume}(a) provides illustrations of how noised volumes appear at various timesteps. Utilizing the standard noise schedule employed by Stable Diffusion, the diffusion process cannot effectively corrupt the information. This is evident as the renderings maintain clarity and recognizability, even at large timesteps. 
Also, it's important to note that there is a huge gap in the information between the noised samples at the final timestep and pure noise. This gap can be perceived as the difference between the training and inference stages. We believe it is due to the high-dimensional character of volume space.

We theoretically analyze the root of this problem. Considering a local patch on the image consisting of $M=w\times h \times c$ values, denoted as $\mathbf{x}_0=\left\{x_0^1,x_0^2,\dots,x_0^M\right\}$. Without loss of generality, we assume that $\{x_0^i\}_{i=1}^M$ are sampled from Gaussian distribution $\mathcal{N}(0,1)$. With common strategy, we add \iid Gaussian noise $\{\epsilon^i\}_{i=1}^M\sim\mathcal{N}(0,1)$ to each value 
by $x_t^i=\sqrt{\gamma_t}x_0^i+\sqrt{1-\gamma_t}\epsilon^i$ to obtain the noised sample, where $\gamma_t$ indicates the noise level at timestep $t$. Thus the expected mean $L_2$ perturbation of the patch is 
\begin{equation}
    \begin{aligned}
        &\ \mathbb{E}  \left (  \frac{1}{M} \sum_{i=0}^M\left (x_0^i -  x_t^i\right ) \right )^2  \nonumber \\
        = &\ \frac{1}{M^2} \mathbb{E} \left ( \sum_{i=0}^M \left (  (1-\sqrt{\gamma_t})\mathbf{x}_0^i -  \sqrt{1-\gamma_t}\epsilon^i \right) \right )^2  \nonumber \\
        = &\ \frac{2}{M}  \left ( 1- \sqrt{\gamma_t}\right ). \nonumber 
    \end{aligned}
\end{equation}

As the resolution $M$ increases, the \iid noises added to each value collectively have a minimal impact on the patch's appearance, and the disturbance is reduced significantly. The rate of information distortion quickly declines to $\frac{1}{M}$. This observation is consistent with findings from concurrent studies~\cite{chen2023importance,gu2022f,hoogeboom2023simple}.
In order to train diffusion models effectively, it's essential to carefully design an appropriate noise that can distort information. So we propose a new noise schedule and the low-frequency noise in the training process.

\subsubsection{New Noise Schedule}

\label{sec:noise_schedule}

The primary goal of our text-to-3D diffusion model is to learn a latent distribution, which is significantly more dimensional than the text-to-image model. As discussed in the previous section, a common noise schedule can lead to insufficient information corruption when applied to high-dimensional spaces, such as volume.

During the training stage, if the information of objects remains a large portion, the network quickly overfits to the noised volumes from the training set and ignores the text conditions. This essentially means that the network leans more towards utilizing information from noised volumes rather than text conditions. To address this, we decided to reduce $\gamma_t$ for all timesteps. Thus we reduced the final signal-to-noise ratio from $6\times 10^{-3}$ to $4\times 10^{-4}$, and evenly reduced $\gamma_t$ at the intermediate timesteps.
Without this, the network may fail to output any object when inference from pure Gaussian noise due to the training and inference gap.

We performed a series of experiments using different noise schedules with various hyper-parameters. These includes the commonly used linear~\cite{ho2020denoising}, cosine~\cite{nichol2021improved}, and sigmoid~\cite{jabri2022scalable} schedules. After comprehensive testing and evaluations, we determined the linear noise schedule to be the most suitable for our experiments.

\subsubsection{Low-Frequency Noise}
Images or feature volumes are typical digital signals, which can be seen as a combination of digital signals of different frequencies.
When adding \iid Gaussian noise to each voxel of a volume, the signal is essentially perturbed by a white noise. The \iid noise evenly corrupts the information of all components through the diffusion process. However, the amplitude of low-frequent components is usually larger and a white noise cannot powerfully corrupt them.
Thus, the mean of the whole volume as the component with the lowest frequency is most likely unnoticed during the diffusion process, causing information leaks. And so are patches and structures of different sizes in the volume. 

Hence, we proposed the low-frequency noise strategy to effectively corrupt information and train diffusion models. We modulate the high-frequency \iid Gaussian noise with an additional low-frequency noise, which is a single value drawn from normal distribution shared by all values in the same channel. Formally, the noise is 

\begin{equation}
    \epsilon^i = \sqrt{1-\alpha}~\epsilon_1^i + \sqrt{\alpha}~\epsilon_2,
    \label{eq:mixed_noise}
\end{equation}

\noindent where $\{\epsilon_1^i\}_{i=1}^M\sim\mathcal{N}(0,1)$ is independently sampled for each location and $\epsilon_2\sim\mathcal{N}(0,1)$ is shared within the patch. We still add noise to data by $x_t^i=\sqrt{\gamma_t}x_0^i+\sqrt{1-\gamma_t}\epsilon^i$, but the noise $\epsilon^i$ is mixed via Equation~\ref{eq:mixed_noise} and no longer \iid.

With the low-frequency noise, the expected mean $L_2$ perturbation of the patch is 
\begin{equation}
    \begin{aligned}
        &\ \mathbb{E}  \left (  \frac{1}{M} \sum_{i=0}^M\left (x_0^i -  x_t^i\right ) \right )^2  \nonumber \\
        = &\ \frac{2}{M}  \left ( 1- \sqrt{\gamma_t}\right ) + (1-\frac{1}{M})(1-\gamma_t)\alpha,\nonumber 
    \end{aligned}
\end{equation}

\noindent where $\alpha\in\left [0,1\right ]$ is a hyper-parameter. The proof is in the supplemental material. By this approach, we introduce additional information corruption that is adjustable and remains scale as the resolution grows, effectively removing information of objects as shown in Figure~\ref{fig:noised_volume}(b).

\subsection{Refinement}

The diffusion model is able to generate a feature volume, but its inherent limitation lies in its output of low-resolution, which restricts texture details. 
To overcome this, we leveraged existing text-to-image models to generate more detailed textures, enhancing the initial results obtained from the diffusion model.

Specifically, we introduced the third stage involving fine-tuning the results. Given the good initial output from the diffusion model, we incorporated SDS~\cite{poole2022dreamfusion} 
in this stage to optimize results, ensuring better image quality and reduced errors. Considering our initial results are already satisfactory, this stage only requires a few iterations, making our entire process still efficient.

Our methodology makes full use of existing text-to-image models to generate textures that are not covered in the original training set, enhancing the details of texture and promoting diversity in the generated images.
Simultaneously, our method also addresses the issue of multiple-face problems encountered in~\cite{poole2022dreamfusion}.

\subsection{Data Filtering}
\label{sec:data_filtering}

We find that data filtering is extremely important to the training. Objaverse is mainly composed of unfiltered user-uploaded 3D models crawled from the web, including many geometry shapes, planer scans and images, texture-less objects, and flawed reconstruction from images. Moreover, the annotation is usually missing or not related, the rotation and position vary in a wide range, and the quality of 3D models is relatively poor compared to image datasets.

Cap3D~\cite{luo2023scalable} propose an approach for automatically generating descriptive text for 3D objects in the Objaverse dataset. They use BLIP-2~\cite{li2023blip}, a pre-trained vision-language model, to caption multi-view rendered images of one object and summarize them into a final caption with GPT-4~\cite{openai2023gpt4}. However, considering the significant variation in captions from different views, even GPT-4 confuses to extract the main concept, hence the final captions are still too noisy for the text-to-3D generation. With these noisy captions, we find that the diffusion model struggles to understand the relation between text conditions and 3D objects.

We generate our own captions with LLaVA~\cite{liu2023llava} and Llama-2~\cite{touvron2023llama} and filter out objects with low-quality or inconsistent multi-view captions in the Objaverse dataset. Similar to Cap3D, we first generate captions of 8 equidistant views around the object and then summarize them into an overall caption with Llama-2. After that, we calculate the similarity matrix of every pair among these 9 captions using CLIP text embedding. We believe that a high-quality 3D object should be visually consistent from different viewpoints, i.e., the captions from different views should be similar. Thus, we use the average and minimal values of the similarity matrix to represent the quality of the object. And manually set two thresholds to filter out objects with low average/minimal similarity scores.

We use a selected subset of objects with the highest quality to train the diffusion model.  We find that the diffusion model is able to learn semantics relations from text conditions. On the contrary, when we use the whole Objaverse dataset for training, the model fails to converge.

\section{Experiments}
\label{sec:experiments}

\subsection{Implementation Details}

\textbf{Dataset} We use the Objaverse~\cite{deitke2023objaverse} dataset in our experiments and rendered $40$ random views for each object.
For the volume encoder, we filter out transparent objects and train with a subset of $750K$ objects. For the diffusion model, we caption and filter as described in Section~\ref{sec:data_filtering} and train with a subset of $100K$ text-object pairs.

\noindent\textbf{Volume encoder} In the first stage, we train a volume encoder that efficiently converts multi-view RGBD images into a feature volume. Each image $\textbf{x}_i$ are fed into a lightweight network $\mathbf{F}$ to extract the feature $\mathbf{F}(\textbf{x}_i)$. The network $\mathbf{F}$ merely includes 2 layers of $5\times 5$ convolution.
Then features of images are unprojected into the coarse volume $\mathbf{v}_c$ and weighted averaged.
$\lambda$ is set to $160N$ in our experiments, where $N=32$ is the spatial resolution of volume. After unprojection, the volume is refined with a 3D U-Net module and rendered with an MLP. The MLP has 5 layers with a hidden dimension of 64. The volume encoder and the rendering decoder in total have $25$M parameters. The model is trained with the Adam~\cite{DBLP:journals/corr/KingmaB14} optimizer. The learning rate is $10^{-4}$ for the volume encoder and $10^{-5}$ for the MLP. The betas are set to $(0.9,0.99)$ and no weight decay or learning rate decay is applied. The input and rendered image resolution is $256^2$ and the batch size of volume is $1$ per GPU. 
We first optimize the model with only $L_2$ loss on the RGB channel. We randomly select $4096$ pixels each from $5$ random views as supervision. After $100 K$ iterations, we add an additional LPIPS loss with a weight of $0.01$. Due to GPU memory limitation, the LPIPS loss is measured on $128\times 128$ patches. The training takes 2 days on 64 V100 GPUs.

\noindent\textbf{Diffusion model} In the second stage, we train a text-conditioned diffusion model to learn the distribution of feature volumes. The denoiser network is a 3D U-Net adopted from~\cite{nichol2022glide}. Text conditions are $77\times 512$ embeddings extracted with CLIP ViT-B/32~\cite{dosovitskiy2020image} text encoder and injected into the 3D U-Net with cross-attentions at middle blocks with spatial resolution $\frac{N}{4}$ and $\frac{N}{8}$. We use a linear~\cite{ho2020denoising} noise schedule with $T=1000$ steps and $\beta_{T}=0.03$. We train with the proposed low-frequency noise strategy and the noise is mixed via Equation~\ref{eq:mixed_noise} with $\alpha=0.5$ in our experiments. The model has $340$M parameters in total and is optimized with the Adam optimizer. The model is supervised by only $L_2$ loss on volumes and no rendering loss is applied. The batch size of volume is $24$ per GPU, the learning rate is $10^{-5}$, the betas are $(0.9,0.99)$, and the weight decay is $2\times 10^{-3}$. The training takes about 2 weeks on 96 V100 GPUs.

\subsection{Volume Encoder}

\begin{figure}[t]
  \centering
   \includegraphics[width=1.0\linewidth]{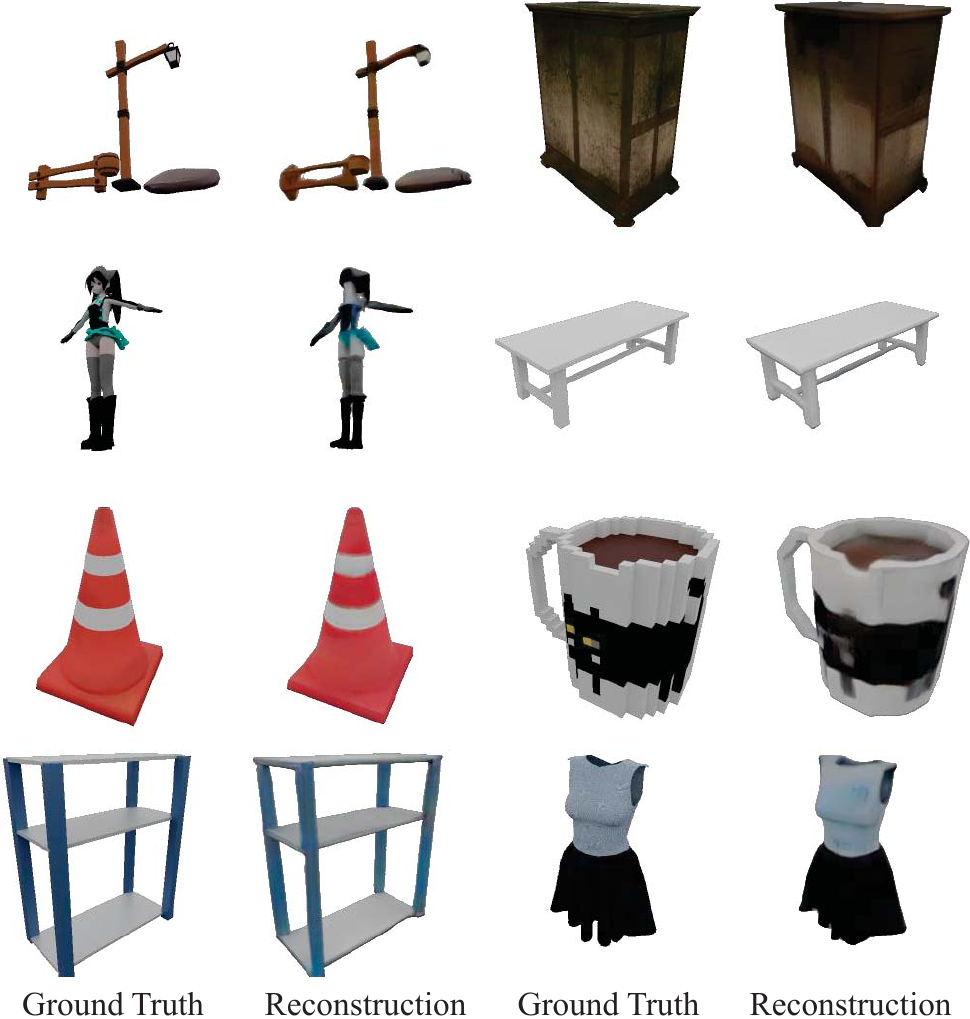}
   \caption{Reconstructions of the volume encoder.}
   \label{fig:stage1_recon}
\end{figure}

\begin{table}[t]
  \centering
  \begin{tabular}{@{}lcccc@{}}
    \toprule
    Data                 & Views & PSNR $\uparrow$  & SSIM  $\uparrow$  & LPIPS $\downarrow$  \\
    \midrule
    \multirow{4}{*}{10K} & 8            & 27.14 & 0.855 & 0.288  \\
                         & 16           & 27.50 & 0.867 & 0.282  \\
                         & 32           & 27.61 & 0.871 & 0.281  \\
                         & 64           & 27.64 & 0.870 & 0.280  \\
    \midrule
    750K (ours)                & 32           & 27.69 & 0.874 & 0.279 \\
    \bottomrule
  \end{tabular}
  \caption{Ablation on input view numbers and training data size of the volume encoder.}
  \label{tab:ablation_encoder}
\end{table}

\begin{figure*}[t]
  \centering
   \includegraphics[width=1.0\linewidth]{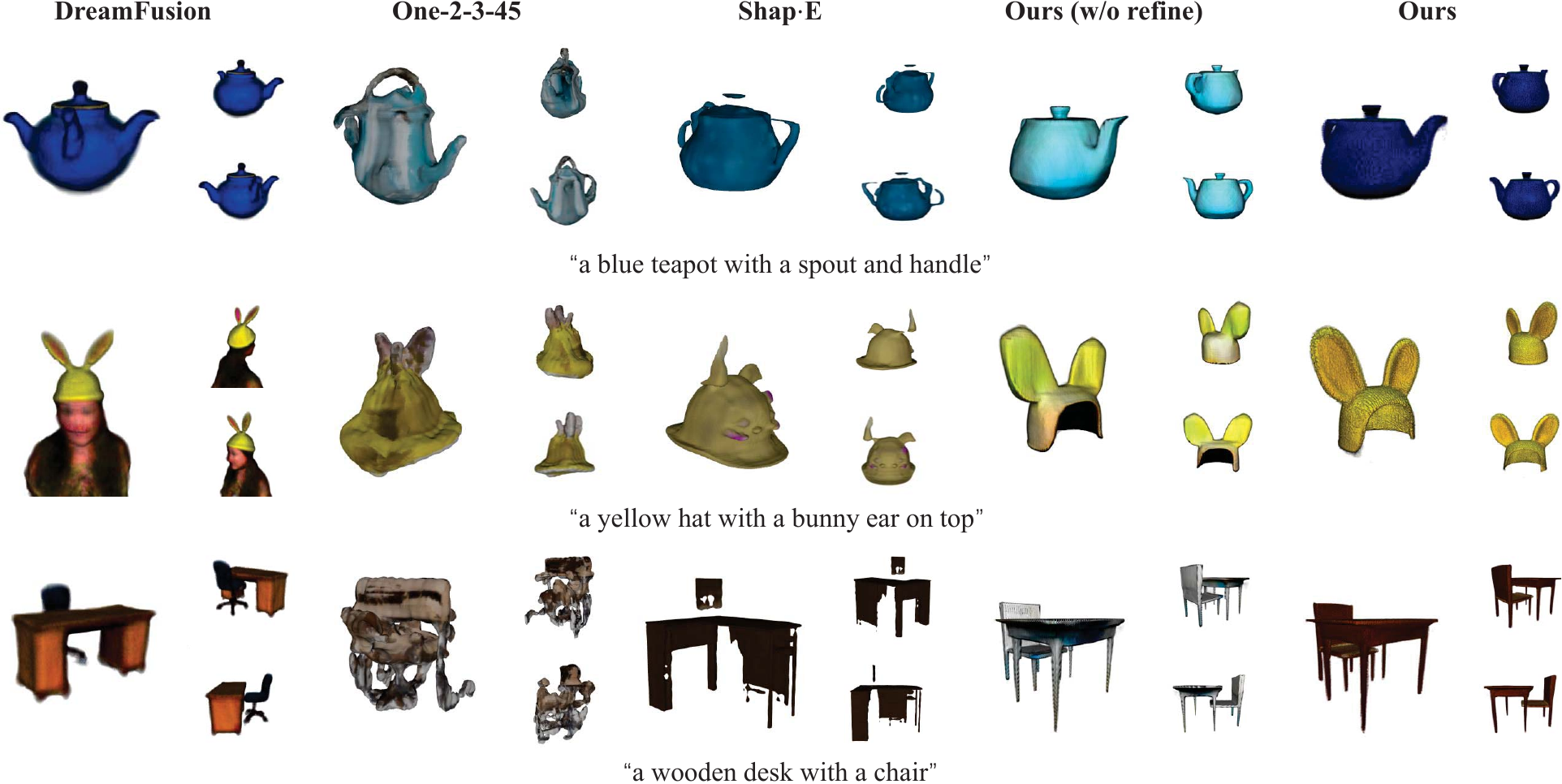}
   \caption{Comparison with state-of-the-art text-to-3D methods.}
   \label{fig:comparison}
\end{figure*}

We first quantitatively study the reconstruction quality of the volume encoder. We set the spatial resolution $N=32$ and channel $C=4$ for efficiency. In Table~\ref{tab:ablation_encoder}, we measure the PSNR, SSIM and LPIPS loss between reconstructions and ground-truth images.
To analyze the correlation between the number of different input views and the quality of reconstruction, we train encoders with different input views on a subset of $10K$ data. It is observed that the quality of reconstruction improves as the number of input views increases. However, once the number of input views surpasses $32$, the enhancement of quality becomes negligible. Therefore, we opted to use $32$ as the default number of input views in our subsequent experiments. Additionally, the quality of reconstruction is also enhanced with the use of more training data.

We show the reconstruction results of the volume encoder in Figure~\ref{fig:stage1_recon}. The volume encoder is capable of reconstructing the geometry shape and textures of objects. Experiments involving higher resolution and larger channels will yield more detailed reconstructions. However, these adjustments will also result in increased training costs and complexity in the second stage. Please refer to the supplemental material for additional ablation studies.

\subsection{Diffusion Model}

\begin{figure*}[t]
  \centering
   \vspace{-0.2cm}
   \includegraphics[width=1.0\linewidth,page=1]{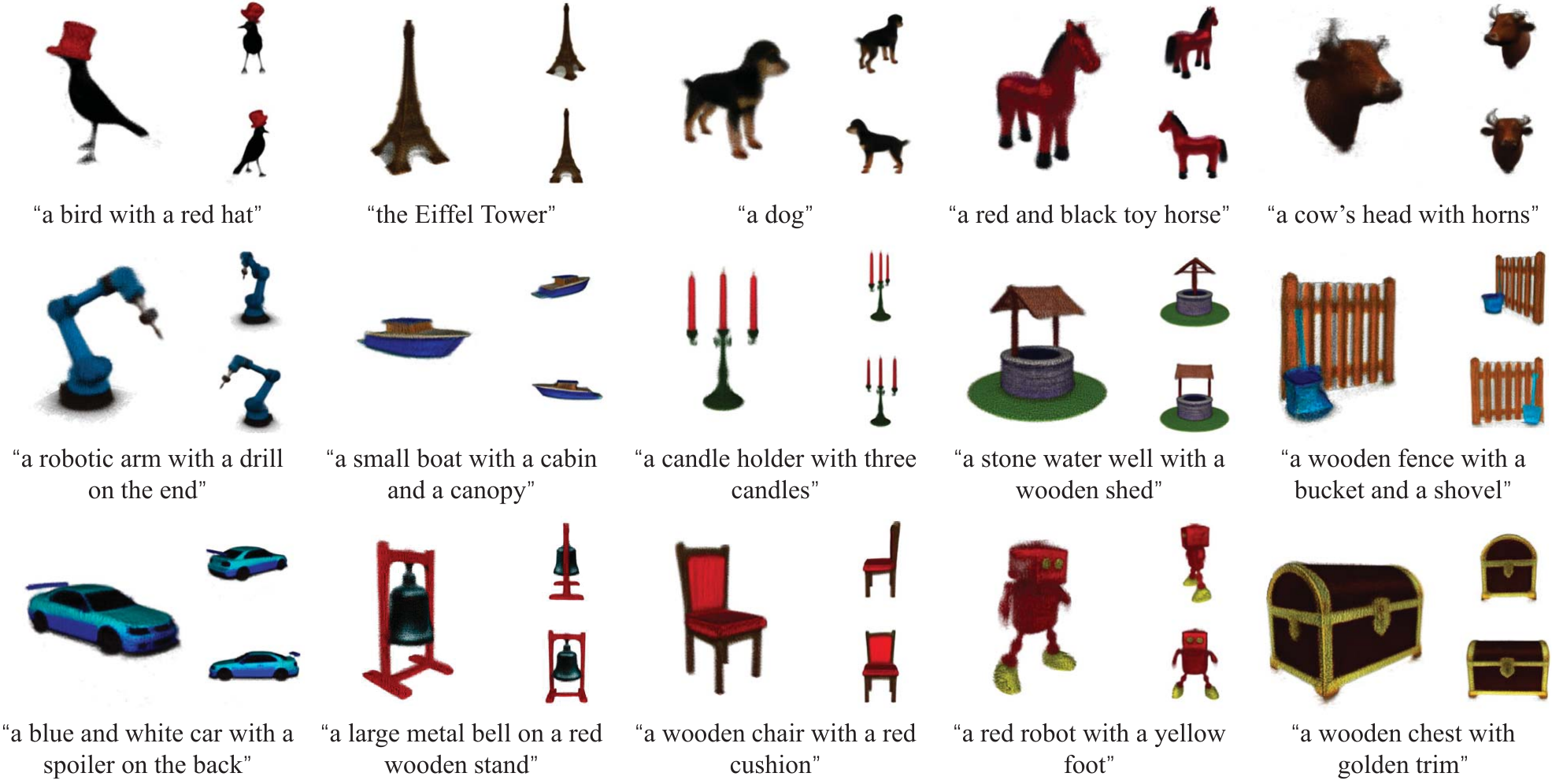}
   \includegraphics[width=1.0\linewidth,page=2]{fig/fig_stage3_inference_results.pdf}
   \caption{Text-to-3D generations by VolumeDiffusion.}
   \label{fig:stage3_inference}
\end{figure*}

We compare our method with state-of-the-art text-to-3D generation approaches, including Shap$\cdot$E~\cite{jun2023shap}, DreamFusion~\cite{poole2022dreamfusion}, and One-2-3-45~\cite{liu2023one}. Since One-2-3-45 is essentially an image-to-3D model, we use images generated with Stable Diffusion as its input. 
Figure~\ref{fig:comparison} demonstrates that our methods yield impressive results, whereas both Shap$\cdot$E and One-2-3-45 struggle to generate complex structures and multiple concepts. For simpler cases, such as a teapot, Shap$\cdot$E, and One-2-3-45 can only produce a rough geometry, with surfaces not so smooth and continuous as those created by our method. For more complex cases, our model excels at combining multiple objects in a scene and aligning better with the text prompts, whereas other methods can only capture parts of the concepts.

Both our method and Shap$\cdot$E are native methods, \ie directly supervised on 3D representation and trained with 3D datasets. It's noteworthy that these native methods generate clearer and more symmetrical shapes (for example, boxes, planes, and spheres) than methods based on image-to-3D reconstruction or distillation. Furthermore, the results of One-2-3-45 are marred by many white dots and stripes, which we believe is due to the inconsistency between images generated by the pre-trained Zero-1-to-3~\cite{liu2023zero} model.

\begin{table}
  \centering
    \begin{tabular}{@{}lcc@{}}
    \toprule
    Method     & Similarity $\uparrow$ & R-Precision $\uparrow$ \\
    \midrule
    DreamFusion~\cite{poole2022dreamfusion} & 0.243      & 47.3\%                 \\
    One-2-3-45~\cite{liu2023one}  & 0.228      & 39.1\%                 \\
    Shap-E~\cite{jun2023shap}      & 0.287      & 58.9\%                 \\
    Ours        & \textbf{0.288}      & \textbf{63.8\%}                 \\
    \bottomrule
  \end{tabular}
  \caption{Quantitative comparison with state-of-the-art text-to-3D methods. Similarity and R-Precision are evaluated with CLIP between rendered images and text prompts.}
  \label{tab:clip_score}
\end{table}

In Table~\ref{tab:clip_score}, we compute the CLIP Similarity and CLIP R-Precision as a quantitative comparison. For each method, we generated 100 objects and rendered 8 views for each object. Our method outperforms others on both visual quality an text alignment.

We present more results in Figure~\ref{fig:stage3_inference}. These prompts include cases of concept combinations and attribute bindings. The critical drawbacks of distillation-based methods, including the Janus problem and over-saturated color, are not observed in our results.

\subsection{Inference Speed}

\begin{table}
  \centering
  \begin{tabular}{@{}llc@{}}
  \toprule
  Stage                   & Method     & Time \\
  \midrule
  \multirow{3}{*}{1 (Encoding)} & Fitting     & $\sim$35min \\
                          & Shap-E~\cite{jun2023shap}      & 1.2sec \\
                          & Ours        & \textbf{33ms} \\
  \midrule
  \multirow{5}{*}{2 (Generation)} & DreamFusion~\cite{poole2022dreamfusion} & $\sim$12hr  \\
                          & One-2-3-45~\cite{liu2023one}  & 45sec      \\
                          & Shap-E~\cite{jun2023shap}      & 14sec      \\
                          & Ours (w/o refine)        & \textbf{5sec}      \\
                          & Ours        & $\sim$5min      \\
  \bottomrule
  \end{tabular}
  \caption{Inference speed comparison. Evaluated on A100 GPU.}
  \label{tab:inference_speed}
   \vspace{-0.2cm}
\end{table}

In Table~\ref{tab:inference_speed}, we report the inference speed of both stages of our method against other approaches. The first stage encodes multi-view images into a 3D representation and is important for scaling up the training data. Shap$\cdot$E uses a transformer-based encoder that takes both $16K$ point clouds and $20$ RGBA images augmented with 3D coordinates as input. It is much slower than our lightweight encoder based on convolution. Fitting means to separately optimize a representation for each object with a fixed rendering MLP, and consumes much more time and storage. The second stage refers to the conditional generation process. Optimization-based DreamFusion needs hours for each object. One-2-3-45, on the other hand, necessitates several diffusion-denoising processes, such as text-to-image and multi-view images generation, and is slower than native 3D methods. For both stages, our method proves to be highly efficient.

\section{Conclusion}
\label{sec:conclusion}

In conclusion, this paper presented a novel method for efficient and flexible generation of 3D objects from text prompts. The proposed lightweight network for the acquisition of feature volumes from multi-view images has been shown to be an efficient method for scaling up the training data required for the diffusion model. The paper also highlighted the challenges posed by high-dimensional feature volumes and presented a new noise schedule and low-frequency noise for improved the training of diffusion models. In experiments, the superior performance of this model in terms of the control of object characteristics through text prompts has been demonstrated. Our future work would focus on refining the algorithm and the network architecture to further speed up the process. We would also involve testing the model on more diverse datasets, including those with more complex objects and varied text prompts.

{
    \small
    \bibliographystyle{ieeenat_fullname}
    \bibliography{main}
}

\clearpage
\setcounter{page}{1}
\maketitlesupplementary

\section{Low-Frequency Noise}

\begin{figure*}
   \centering
   \includegraphics[width=0.9\textwidth]{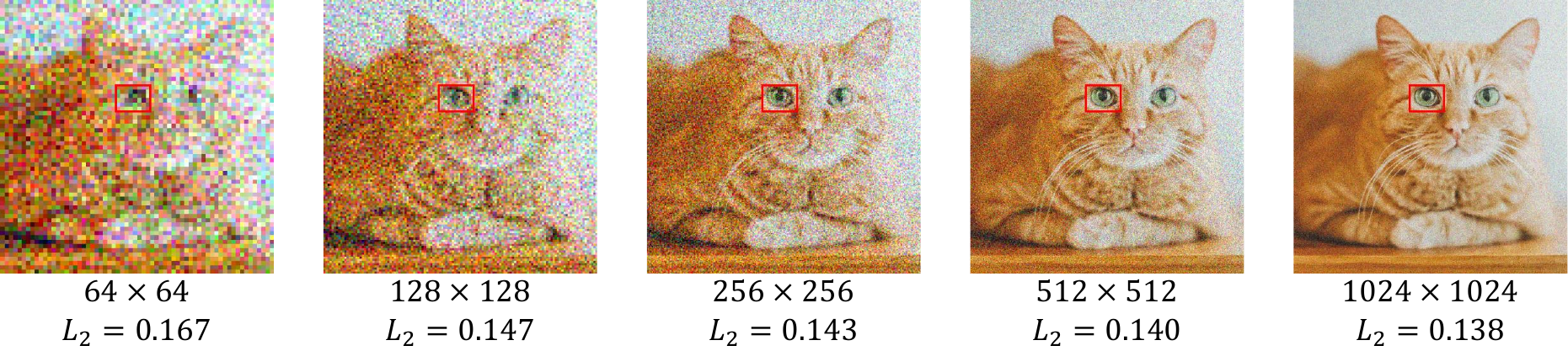}
   \caption{Noised images with different resolutions. All images are noised with $x_t=\sqrt{\gamma}x_0+\sqrt{1-\gamma}\epsilon$ and $\gamma=0.65$, $\epsilon\sim\mathcal{N}(0,1)$.}
   \label{fig:supp_motivation}
\end{figure*}

\subsection{Formula derivation}

In this section, we present a detailed derivation of the expected mean $L_2$ perturbation of a patch in Section~\ref{sec:diffusion_model}.

Consider a patch $\mathbf{x}_0=\left\{x_0^1,x_0^2,\dots,x_0^M\right\}$. We add noise $\{\epsilon^i\}_{i=1}^M$ to each value 
by $x_t^i=\sqrt{\gamma_t}x_0^i+\sqrt{1-\gamma_t}\epsilon^i$ to obtain the noised sample, where $\gamma_t$ indicates the noise level at timestep $t$. The expected mean $L_2$ perturbation of the patch $\mathbf{x}_0$ with \iid Gaussian noise $\{\epsilon^i\}_{i=1}^M\sim\mathcal{N}(0,1)$ is

\begin{equation}
    \begin{aligned}
        &\ \mathbb{E}  \left (  \frac{1}{M} \sum_{i=0}^M\left (x_0^i -  x_t^i\right ) \right )^2  \nonumber \\
        = &\ \frac{1}{M^2} \mathbb{E} \left ( \sum_{i=0}^M \left (  (1-\sqrt{\gamma_t})x_0^i -  \sqrt{1-\gamma_t}\epsilon^i \right) \right )^2  \nonumber \\
        = &\ \frac{(1-\sqrt{\gamma_t})^2}{M^2} \mathbb{E} \left( \sum_{i=0}^M x_0^i\right)^2 +  \frac{1-\gamma_t}{M^2} \mathbb{E} \left( \sum_{i=0}^M \epsilon^i \right )^2 \\ &\ - \frac{(1-\sqrt{\gamma_t})\sqrt{{1-\gamma_t}}}{M^2} \mathbb{E} \left( \sum_{\substack{i=0 \\ j=0}}^M x_0^i\epsilon^j\right) \nonumber \\
        = &\ \frac{(1-\sqrt{\gamma_t})^2}{M^2} \left( \mathbb{E} \sum_{i=0}^M \left(x_0^i\right)^2+ \mathbb{E} \sum_{i\ne j}^M \left(x_0^i x_0^j\right)\right) \\
          &\ + \frac{1-\gamma_t}{M^2} \left( \textcolor{blue}{ \mathbb{E} \sum_{i=0}^M \left(\epsilon^i\right)^2 } + \textcolor{red}{ \mathbb{E} \sum_{i\ne j}^M \left(\epsilon^i \epsilon^j\right) } \right) \nonumber \\
        = &\ \frac{1}{M}(1-\sqrt{\gamma_t})^2 + \frac{1}{M}(1-\gamma_t) \\
        = &\ \frac{2}{M}  \left ( 1- \sqrt{\gamma_t}\right ). \nonumber 
    \end{aligned}
\end{equation}

With the proposed low-frequency noise strategy, we mix the noise by $\epsilon^i = \sqrt{1-\alpha}~\epsilon_1^i + \sqrt{\alpha}~\epsilon_2$ (Equation~\ref{eq:mixed_noise}), where $\{\epsilon_1^i\}_{i=1}^M\sim\mathcal{N}(0,1)$ is independently sampled for each location and $\epsilon_2\sim\mathcal{N}(0,1)$ is shared within the patch. We still add noise by $x_t^i=\sqrt{\gamma_t}x_0^i+\sqrt{1-\gamma_t}\epsilon^i$ and only $\epsilon^i$ is changed. So we have 

\begin{equation}
    \begin{aligned}
        &\  \textcolor{blue}{ \mathbb{E} \sum_{i=0}^M \left(\epsilon^i\right)^2 } \nonumber \\
        = &\  \mathbb{E} \sum_{i=0}^M \left((1-\alpha)(\epsilon_1^i)^2+\alpha(\epsilon_2)^2+2\sqrt{\alpha(1-\alpha)}\epsilon_1^i\epsilon_2\right) \nonumber \\
        = &\ (1-\alpha) \mathbb{E} \sum_{i=0}^M (\epsilon_1^i)^2 + \alpha \mathbb{E} \sum_{i=0}^M (\epsilon_2)^2 \nonumber \\
        = &\ M,
    \end{aligned}
\end{equation}

\begin{equation}
    \begin{aligned}
        &\ \textcolor{red}{ \mathbb{E} \sum_{i\ne j}^M \left(\epsilon^i\epsilon^j\right) } \nonumber \\
        = &\ \mathbb{E} \sum_{i\ne j}^M \left((\sqrt{1-\alpha}\epsilon_1^i+\sqrt{\alpha}\epsilon_2)(\sqrt{1-\alpha}\epsilon_1^j+\sqrt{\alpha}\epsilon_2)\right) \nonumber \\ 
        = &\ \mathbb{E} \sum_{i\ne j}^M \left((1-\alpha)\epsilon_1^i\epsilon_1^j+\alpha(\epsilon_2)^2\right) \nonumber \\ 
        = &\ \alpha \sum_{i\ne j}^M \mathbb{E} (\epsilon_2)^2 \nonumber \\
        = &\ \alpha M(M-1).
    \end{aligned}
\end{equation}

\noindent In conclusion, the expected mean $L_2$ perturbation of the patch $\mathbf{x}_0$ with the low-freqency noise is

\begin{equation}
    \begin{aligned}
        &\ \mathbb{E}  \left (  \frac{1}{M} \sum_{i=0}^M\left (x_0^i -  x_t^i\right ) \right )^2  \nonumber \\
        = &\ \frac{1}{M}(1-\sqrt{\gamma_t})^2 + \frac{1-\gamma_t}{M^2}\left(M+\alpha M(M-1)\right) \\
        = &\ \frac{2}{M}  \left ( 1- \sqrt{\gamma_t}\right ) + (1-\frac{1}{M})(1-\gamma_t)\alpha. \nonumber
    \end{aligned}
\end{equation}

\begin{figure}[t]
  \centering
   \includegraphics[width=0.9\linewidth]{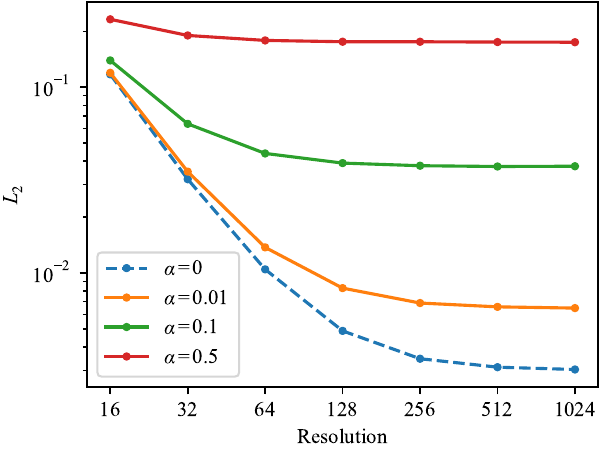}
   \caption{Patch $L_2$ perturbation of noised images at timestep $t=200$. $\alpha=0$ refers to \iid noise. As image resolution increases, the $L_2$ distortion with our proposed noise is almost unaffected and remains at a high level.}
   \label{fig:patch_l2}
\end{figure}

Here we assume that $\{x_0^i\}_{i=1}^M$ are also sampled from Gaussian distribution $\mathcal{N}(0,1)$, which may be not true on real data. Thus we report the mean $L_2$ perturbations on real images with different resolutions in Figure~\ref{fig:patch_l2} as a further demonstration. As illustrated, the $L_2$ perturbation of \iid noise decays exponentially as resolution increases, while our proposed low-frequency noise is slightly affected and converges to larger values proportional to $\alpha$.

\subsection{Justification for patchwise mean $L_2$ loss}

To validate the reasonableness of our adoption of patchwise mean $L_2$ perturbation, we follow \cite{chen2023importance} and present an intuitive example using 2D images in Figure~\ref{fig:supp_motivation}. The red rectangle highlights the same portion of the object across different resolutions, and we calculate the patchwise $L_2$ loss for each. We observe that as the image resolution increases, the loss diminishes even though these images maintain the same noise level ($\gamma=0.65$), making the denoising task easier for networks. Consequently, we believe it is essential to reassess noises from the local patch perspectives and propose the expected mean $L_2$ perturbation of a patch as a metric.

\subsection{Can adjusting the noise schedule also resolve the issue in Figure~\ref{fig:noised_volume}?}

In relation to the issue of incomplete removal information in Figure~\ref{fig:noised_volume} of the main paper, we rely on the low-frequency noise schedule to solve it. However, the question arises: can this issue also be addressed solely by adjusting the noise schedule as mentioned in Section~\ref{sec:noise_schedule}?

The answer is negative. Let's consider a scenario where we modify the noise schedules $\gamma_t$ and $\gamma_t'$ for spaces with resolution $M$ and $M'$ respectively, ensuring that the $L_2$ perturbation remains constant:

\begin{equation}
    \begin{aligned}
       &\ \frac{2}{M'} \left( 1 - \sqrt{\gamma_t'} \right) = \frac{2}{M} \left( 1 - \sqrt{\gamma_t} \right)  \\
       \Leftrightarrow &\ \frac{1 - \sqrt{\gamma_t'}}{1 - \sqrt{\gamma_t}} = \frac{M'}{M}  \label{eq:noise}\\
    \end{aligned}
\end{equation}

We take the default setting in Stable Diffusion, where $\beta_T=0.012$ as an example, leading to $\gamma_T=0.048$. The volumn resolution (where $M'=32^3$) is $8$ times larger than default resolution ($M=64^2$). Substituting these values into Equation~\ref{eq:noise}, we find that there is no solution for $\gamma_t'$. This suggests that adjusting noise schedule alone is not a viable solution for high-dimensional spaces.

\subsection{Ablation}

\begin{table}[t]
  \centering
    \begin{tabular}{@{}llcc@{}}
    \toprule
    $\alpha$ & $\beta_T$ & Similarity $\uparrow$ & R-Precision $\uparrow$ \\
    \midrule
    0 & 0.02 & 0.198 & 11.3\% \\
    0 & 0.03 & 0.264 & 50.7\% \\
    0.5 & 0.02 & 0.201 & 11.7\% \\
    0.5 & 0.03 & \textbf{0.279} & \textbf{56.5\%} \\
    \bottomrule
  \end{tabular}
  \caption{Quantitative comparison between models trained with different noise strategy. $\alpha=0$ refers to \iid noise.}
  \label{tab:noise_ablation}
\end{table}

\begin{table*}[t]
  \centering
  \begin{tabular}{@{}c|cc|ccc@{}}
    \toprule
     & Resolution $N$ & Channel $C$  & PSNR $\uparrow$ & SSIM $\uparrow$ & LPIPS $\downarrow$ \\
    \midrule
    exp01 & 32 & 32 &  27.83 & 0.874 & 0.276 \\
    exp02 & 32 & 4 &  27.69 & 0.874 & 0.279 \\
    exp03 & 64 & 4 &  29.21 & 0.886 & 0.228 \\
    \midrule
    exp04 & 32 $\rightarrow$ 64 & 4 & 28.68 & 0.883 & 0.167 \\
    \bottomrule
  \end{tabular}
  \caption{Ablation experiments on the volume encoder.}
  \label{tab:ablation_lpips}
\end{table*}

We conducted ablation experiments on noise schedule and the low-frequency noise in Table~\ref{tab:noise_ablation}. We trained diffusion models with $a=\{0, 0.5\}$ and $\beta_T=\{0.02,0.03\}$ on a subset of $5K$ data and compares CLIP Similarity and R-Precision. The results demonstrate the effectiveness of our noise strategy.

On noise schedule, we find $\beta_T=0.02$ performs poorly, as the models fail to output any objects while inferencing from pure Gaussian noise. We believe it is due to the information gap between the last timestep and pure noise, which is illustrated in Figure~\ref{fig:noised_volume}(a). Meanwhile, models trained with $\beta_T=0.03$ eliminate the training-inference gap and are able to draw valid samples from pure noise.

On noise types, we find the model trained with \iid noise ($\alpha=0$) has lower scores, as it tends to exploit the remaining information of noised volume and confuses when starting from Gaussian noise. In the contrary, the model trained with the low-frequency noise ($\alpha=0.5$) is forced to learn from text conditions and produces results that are more preferable and consistent to text prompts.

\section{Unprojection}

The volume encoder takes a set of input views $(\textbf{x}, \textbf{d}, \textbf{p})$, where $\textbf{x} = \{x^{(i)}\in\mathbb{R}^{3\times H\times W}\}_{i=1}^{N}$ are images, $\textbf{d} = \{d^{(i)}\in\mathbb{R}^{H\times W}\}_{i=1}^{N}$ are corresponding depths and $\textbf{p} = \{p^{(i)}\in\mathbb{R}^{4\times 4}\}_{i=1}^{N}$ are camera poses. The camera pose $p^{(i)}$ can be explicitly written as

\begin{equation}
    p^{(i)} = \begin{bmatrix}
                R^{(i)} & t^{(i)} \\ 
                0 & 1
            \end{bmatrix},
\end{equation}

\noindent where $R^{(i)}\in\mathbb{R}^{3\times 3}$ is the camera rotation and $t^{(i)}\in\mathbb{R}^{3}$ is the camera position.

We we obtain the coarse volume $\mathbf{v}_c$ by the unprojection

\begin{equation}
   \mathbf{v}_c=\Phi(\mathbf{F}(\textbf{x}),\textbf{d},\textbf{p}),
\end{equation}

\noindent where $\mathbf{F}(\cdot)$ is the feature extractor network and $\Phi(\cdot)$ is the unprojection operation.

We first set up an auxiliary coordinate volume $V_{coord}=\{(x_i,y_i,z_i,1)\}$, where $x_i,y_i,z_i\in[-1,1]$ is the coordinate of the $i$-th voxels of $\mathbf{v}_c^i$ in space. We project the 3D space coordinate $V_{coord}^j=(x_j,y_j,z_j,1)$ of the $j$-th voxel into 2D space of the $i$-th image by

\begin{equation}
    X_{coord}^{i,j} = \kappa \cdot \left(p^{(i)}\right)^{-1} \cdot V_{coord}^j,
\end{equation}

\noindent where $\kappa\in\mathbb{R}^{3\times 4}$ is the camera intrinsic, \eg focal length, and $X_{coord}^{i,j}=\{u_{i,j},v_{i,j},w_{i,j}\}$. $\frac{1}{w_{i,j}}X_{coord}^{i,j}$ is the coordinate in the 2D space defined by the $i$-th image.

Then we perform sampling with

\begin{equation}
    f_{i,j} = \phi\left(\frac{1}{w_{i,j}}X_{coord}^{i,j}, \mathbf{F}(x^{(i)})\right)
\end{equation}

\noindent where $\phi(x,y)$ is the grid sampling function that samples value from $y$ according to the coordinate $x$. We also sample the ground-truth depth by 

\begin{equation}
    d_{i,j} = \phi\left(\frac{1}{w_{i,j}}X_{coord}^{i,j}, d^{(i)}\right).
\end{equation}

Finally, we aggregate features from different views with the weighted average

\begin{equation}
    \mathbf{v}_c^j=\frac{1}{\sum_{i=0}^N w_{i,j}}\left(\sum_{i=0}^N w_{i,j}f_{i,j}\right).
\end{equation}

\noindent The weight $w_{i,j}$ is obtained by applying a Gaussian function on the depth difference $\Delta d_{i,j}$

\begin{equation}
    w_{i,j} = \exp\left(-\lambda(\Delta d_{i,j})^2\right),
\end{equation}

\noindent where $\Delta d_{i,j}=d_{i,j}-\hat{d}_{i,j}$ and $\hat{d}_{i,j}=\left\|V_{coord}^j-t^{(i)}\right\|_2$ is the calculated distance between the $j$-th voxel and the camera of the $i$-th image.

\section{Volume Encoder}

In Table~\ref{tab:ablation_lpips}, we conducted ablation experiments to study how resolution $N$, channel $C$ and loss term affects the performance of the volume encoder.

We find the channel $C$ of volume is a minor factor of the performance of the volume encoder. In contrast, increasing the resolution $N$ greatly improves the reconstruction performance. However, $N=64$ brings a computation and GPU memory cost that is $8$ times larger than $N=32$, which causes significant difficulty for training diffusion models.

In order to increase volume resolution without large overhead, we introduce a super-resolution module before we feed the generated volume into the refinement module. We increase the spatial resolution of the volume from $N=32$ to $N=64$. The super-resolution module composed of few layers of 3D convolution is served as a post-process and is performed on the outputs of the diffusion model. In our experiments, the super-resolution approach achieves close performances comparing to native $N=64$ volumes. The diffusion model is trained on the volumes with lower resolution $N=32$, and the rendering is performed on the upsampled volumes with higher resolution $N=64$. Therefore, we can enjoy both a lower dimension for easier training of diffusion models as well as a higher resolution for rendering more detailed textures without much overhead.

\section{Limitation}

\begin{figure}[t]
  \centering
   \includegraphics[width=0.88\linewidth]{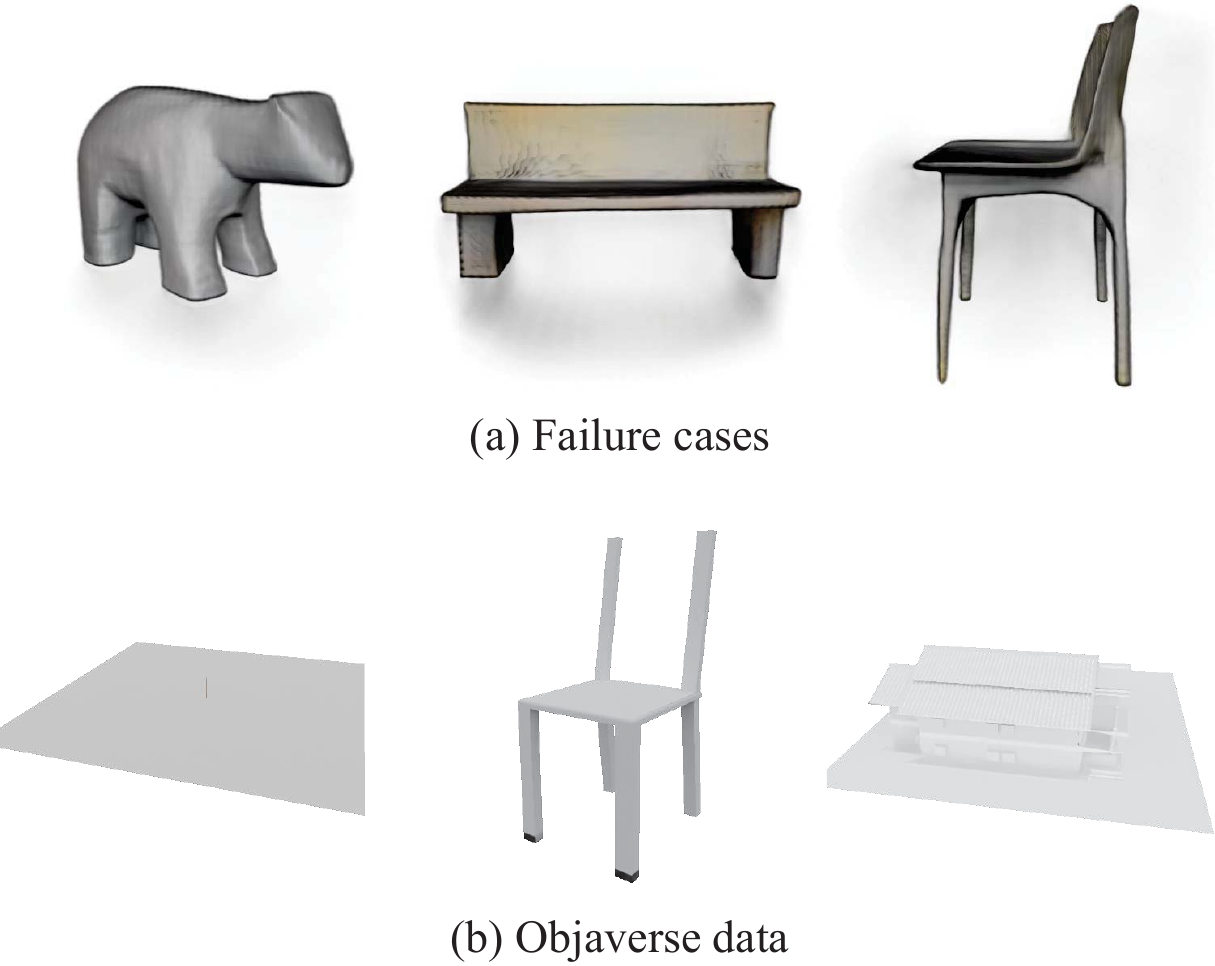}
   \caption{Limitations of the proposed method.}
   \label{fig:failure}
\end{figure}

Our method has two main drawbacks and we present three typical failure cases in Figure~\ref{fig:failure}. 

First, both the volume encoder and the diffusion model are trained on Objaverse~\cite{deitke2023objaverse} dataset. However, the dataset includes many white objects with no texture as illustrated. As a consequence, our model usually prioritizes geometry over color and texture, and is biased towards generating white objects.

Second, the 3D objects generated by our model usually have over-smooth surfaces and shapes. We believe this is attributed to the relatively low spatial resolution $N=32$ of feature volumes. However, with a higher resolution $N=64$, the dimension of the latent space is $8$ times larger and the diffusion model struggles to converge. Due to the GPU resources limit, we will leave it to our future works.

\section{More results}

\begin{figure*}[t]
  \centering
   \includegraphics[width=1.0\linewidth,page=1]{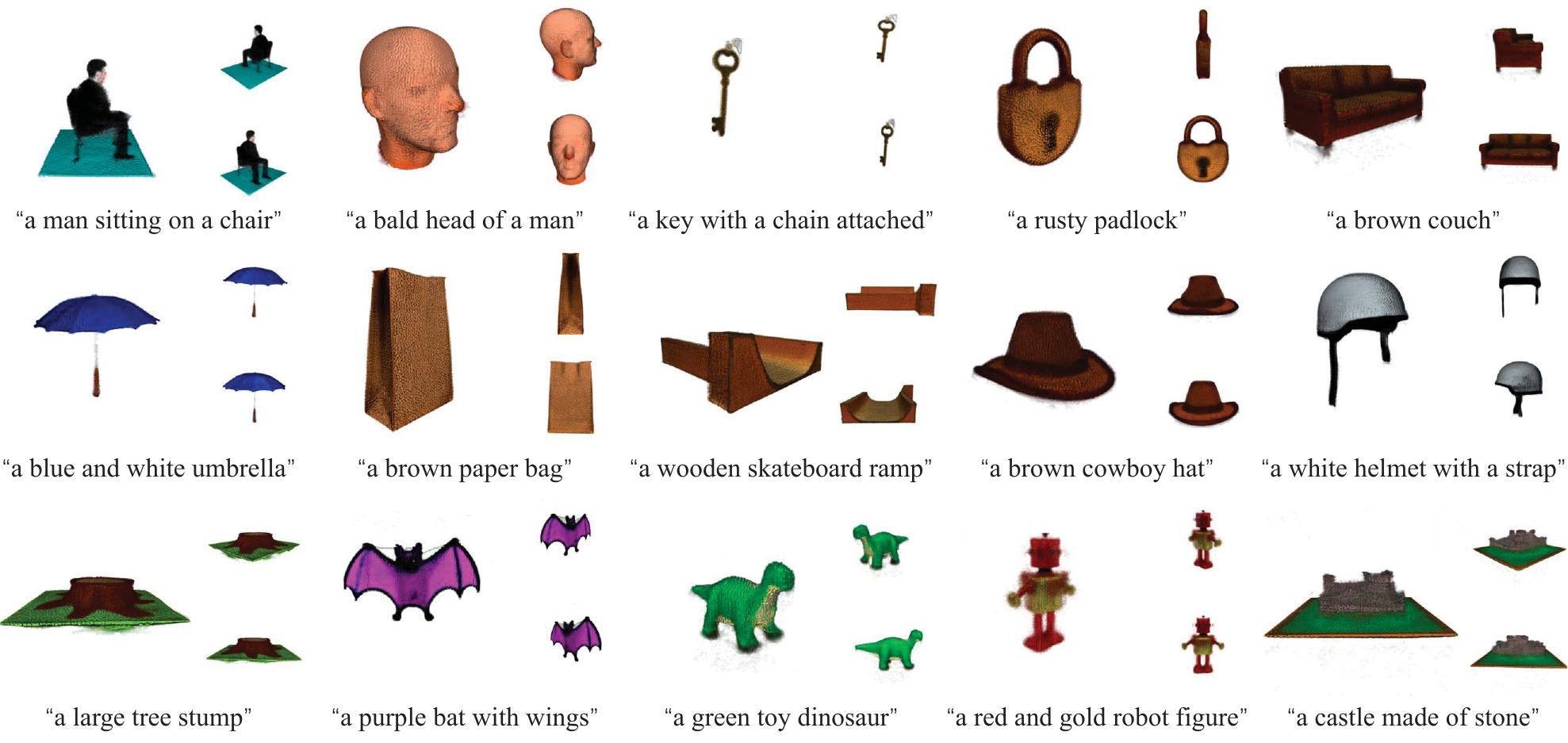}
   \includegraphics[width=1.0\linewidth,page=2]{fig/fig_supplementary.pdf}
   \caption{More text-to-3D generations of VolumeDiffusion.}
   \label{fig:supp_results_1}
\end{figure*}

\begin{figure*}[t]
  \centering
   \includegraphics[width=1.0\linewidth,page=3]{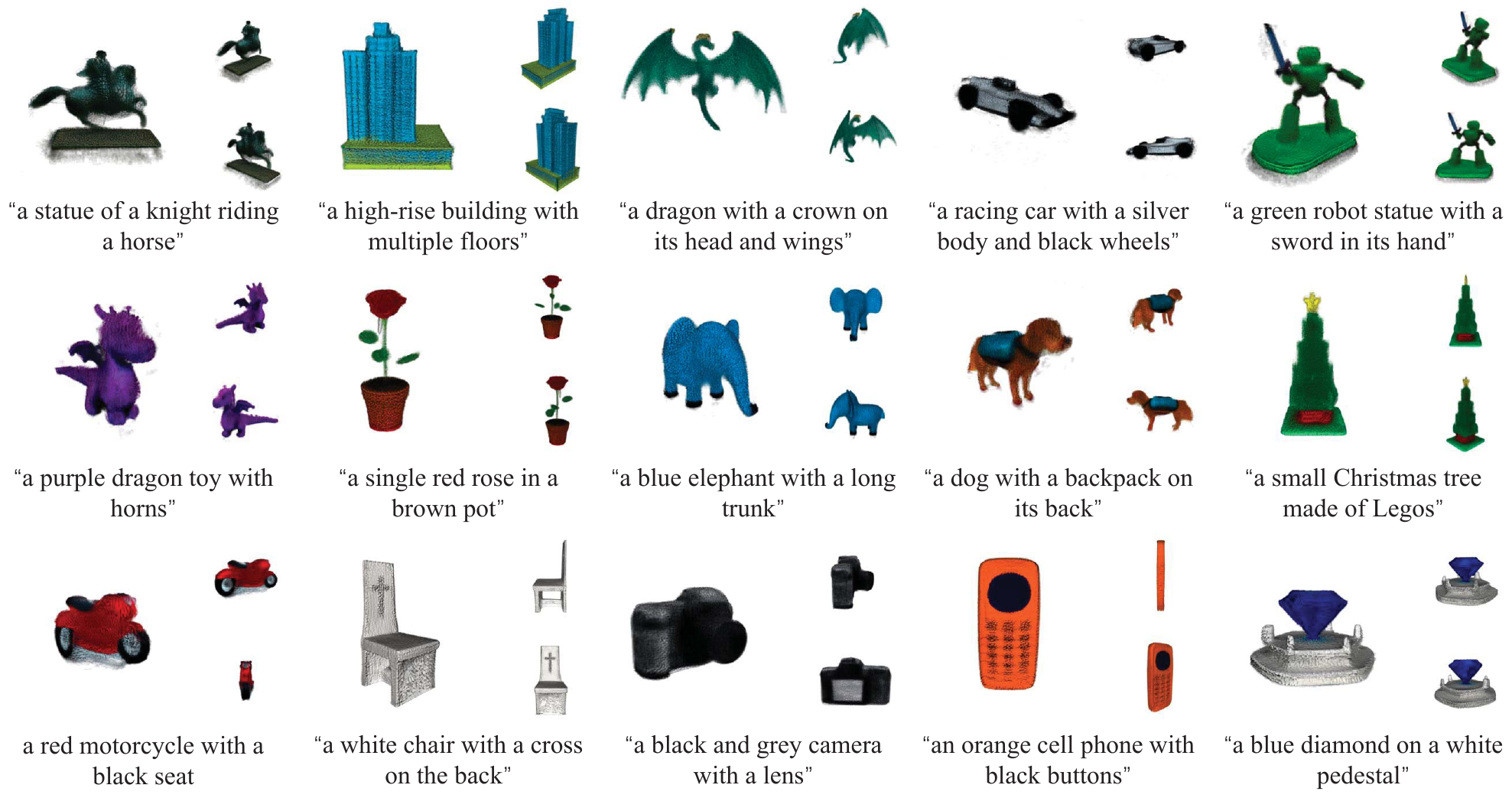}
   \includegraphics[width=1.0\linewidth,page=4]{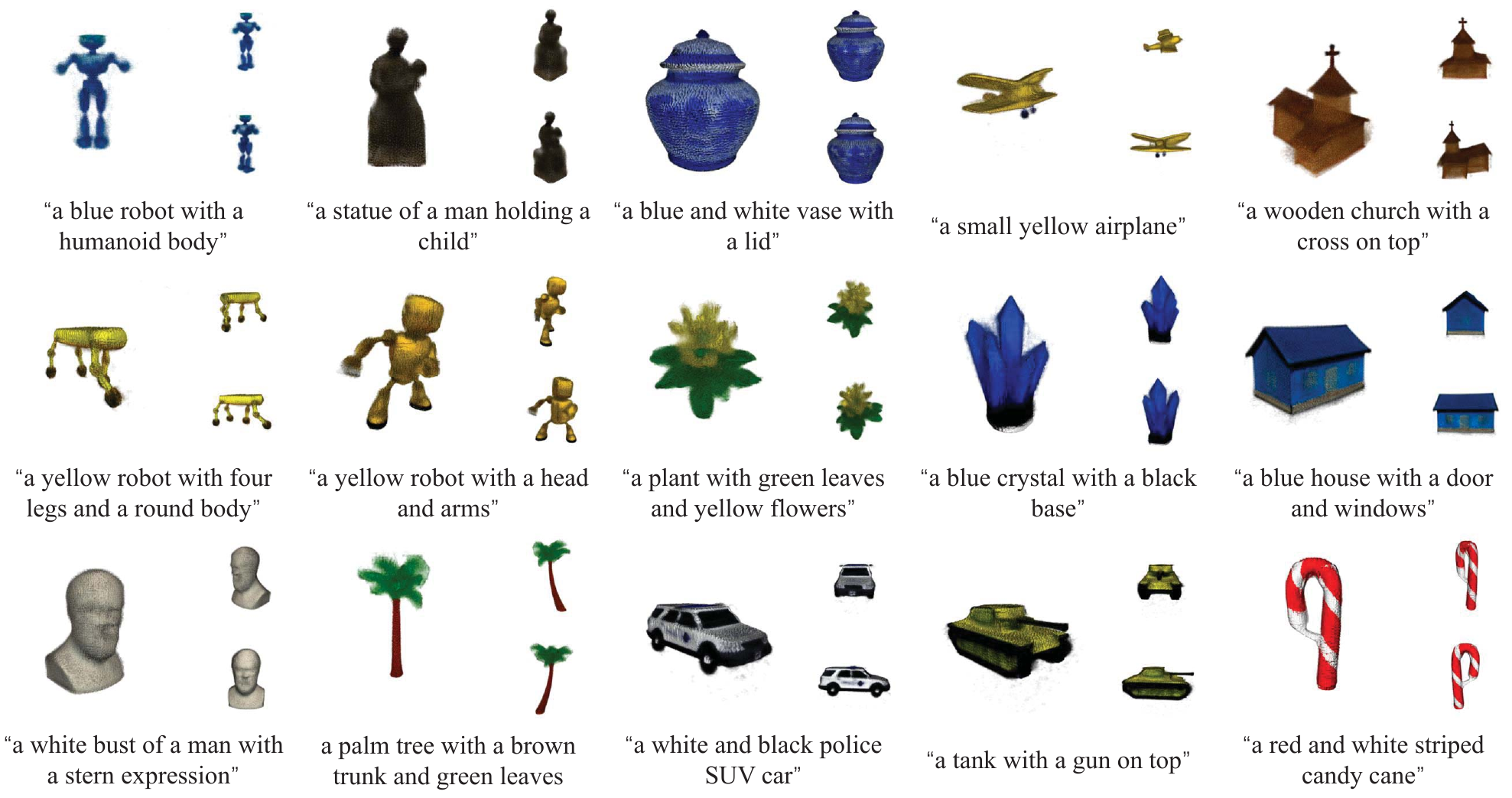}
   \caption{More text-to-3D generations of VolumeDiffusion.}
   \label{fig:supp_results_2}
\end{figure*}

\begin{figure*}[t]
  \centering
   \includegraphics[width=1.0\linewidth,page=5]{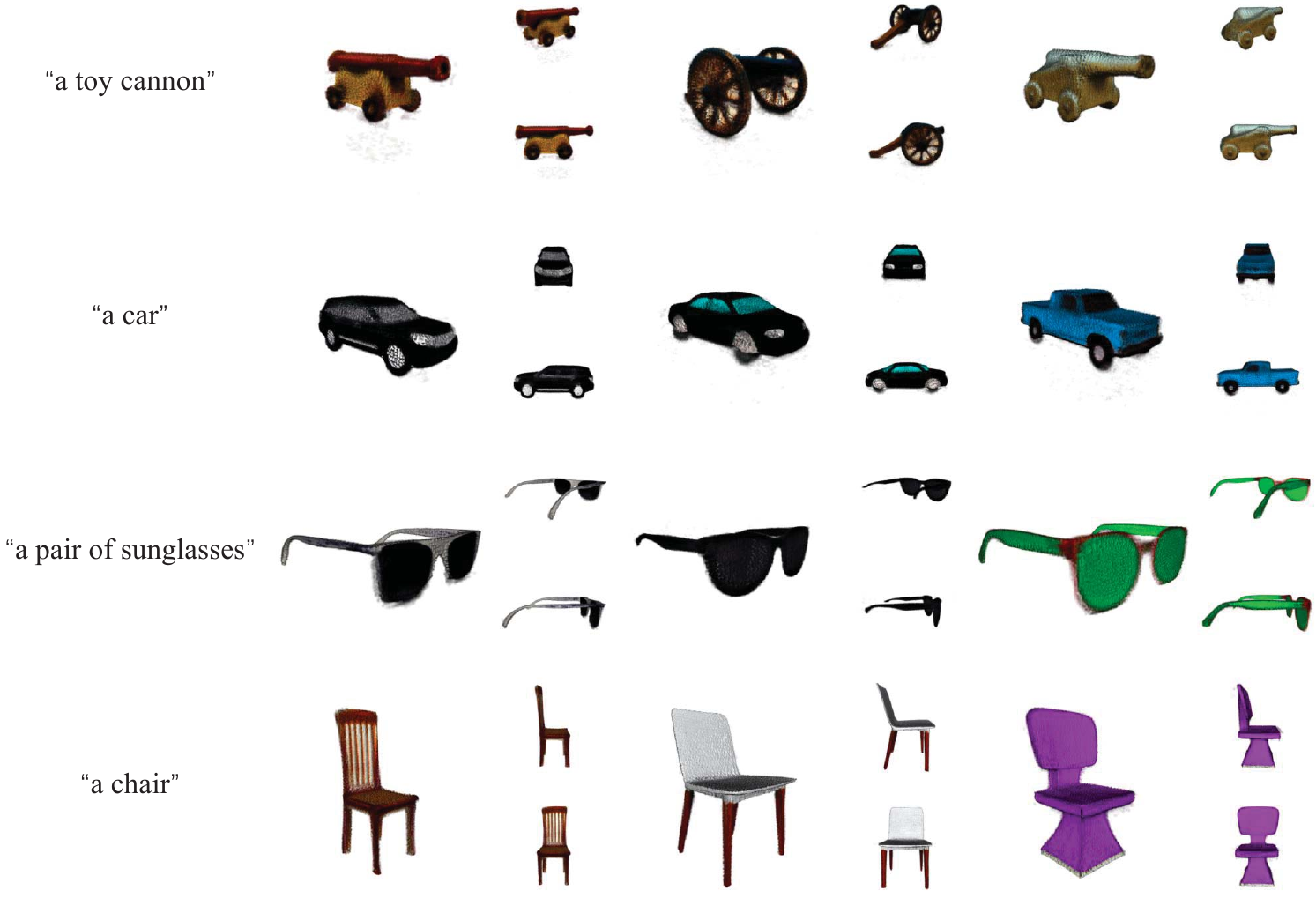}
   \caption{Diverse text-to-3D generations of VolumeDiffusion.}
   \label{fig:supp_diversity}
\end{figure*}

\begin{figure*}[t]
  \centering
   \includegraphics[width=1.0\linewidth,page=6]{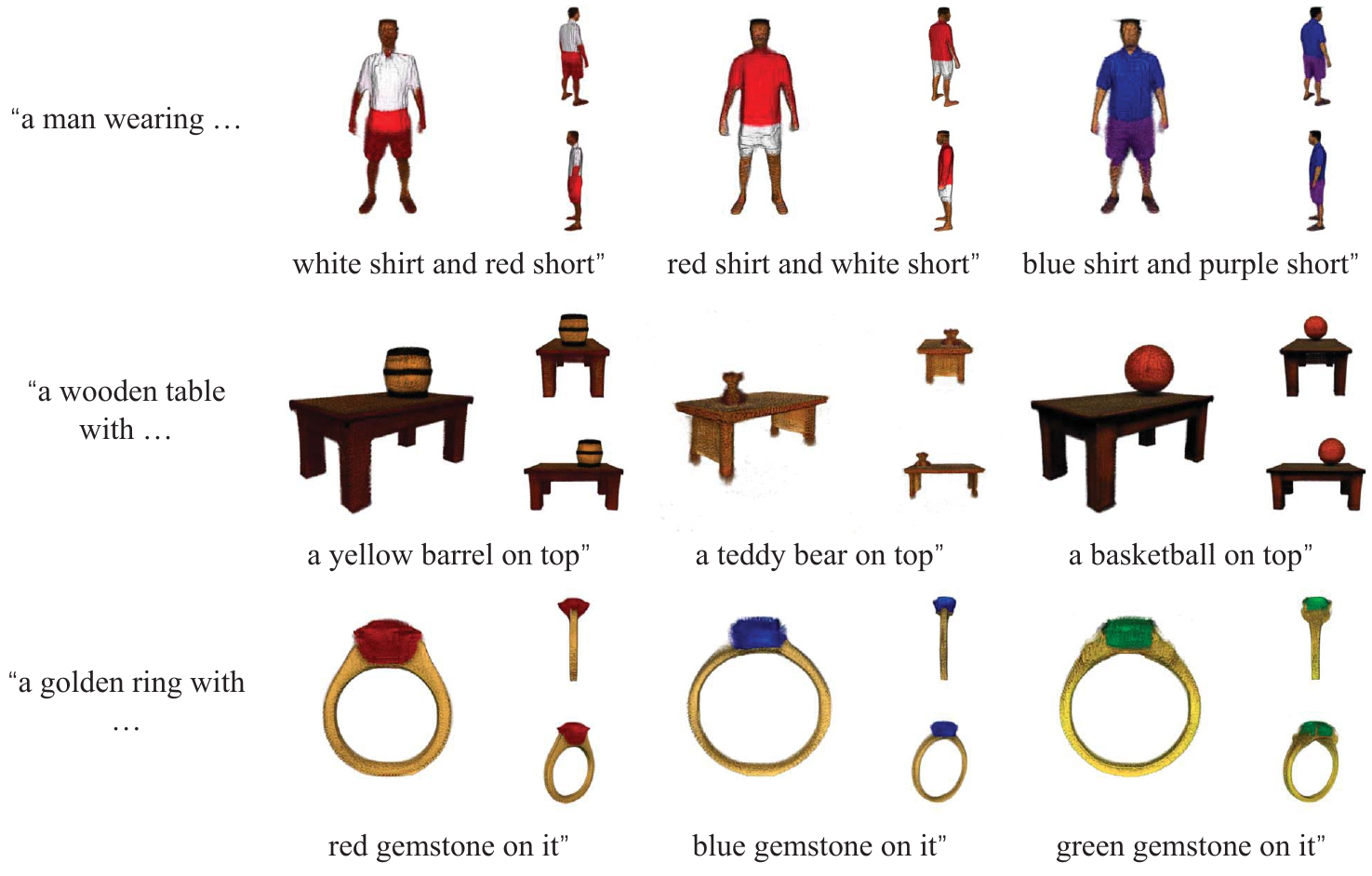}
   \caption{Flexible text-to-3D generations of VolumeDiffusion.}
   \label{fig:supp_flexibility}
\end{figure*}

\begin{figure*}[t]
  \centering
   \includegraphics[width=1.0\linewidth,page=7]{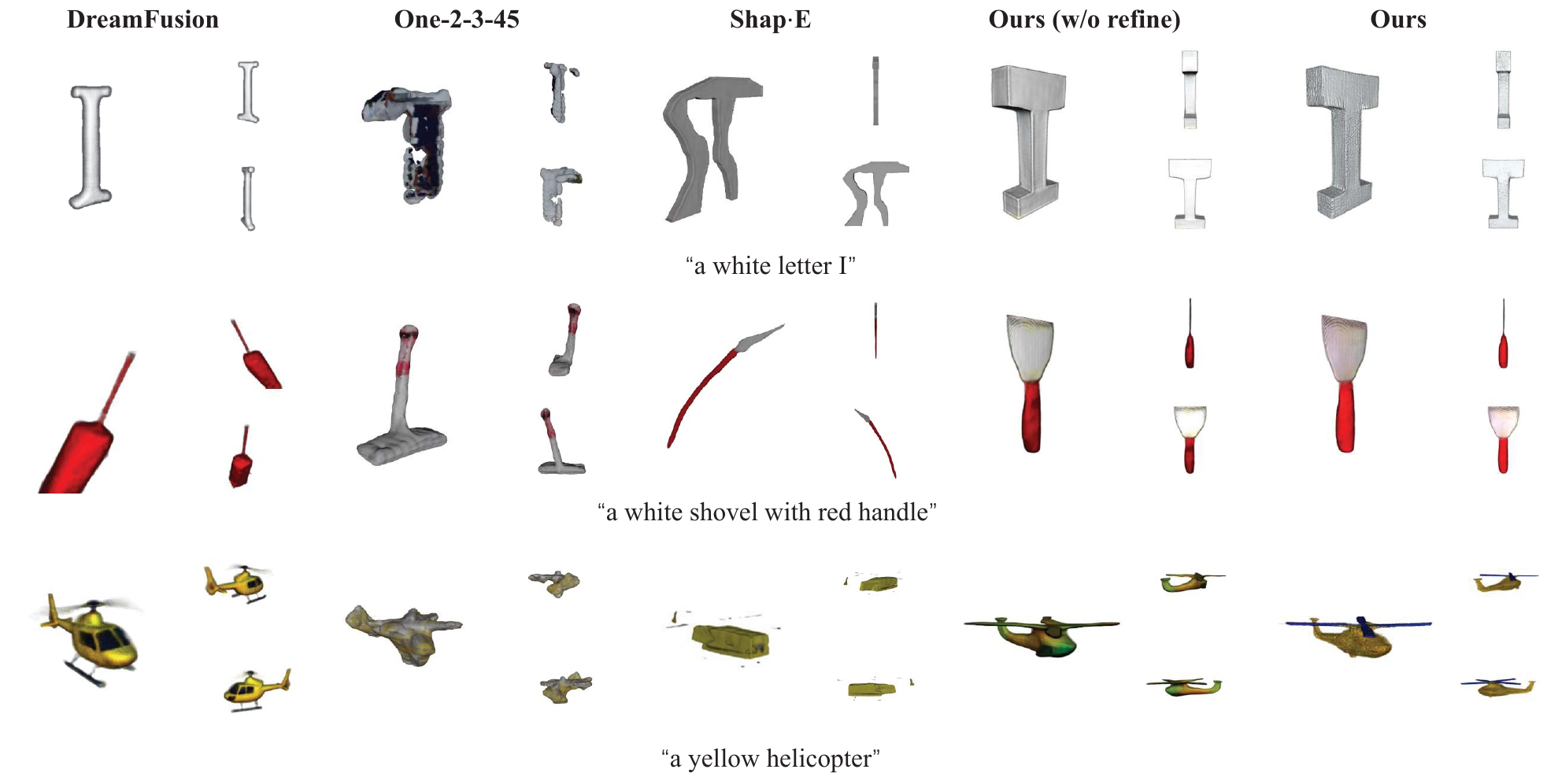}
   \includegraphics[width=1.0\linewidth,page=8]{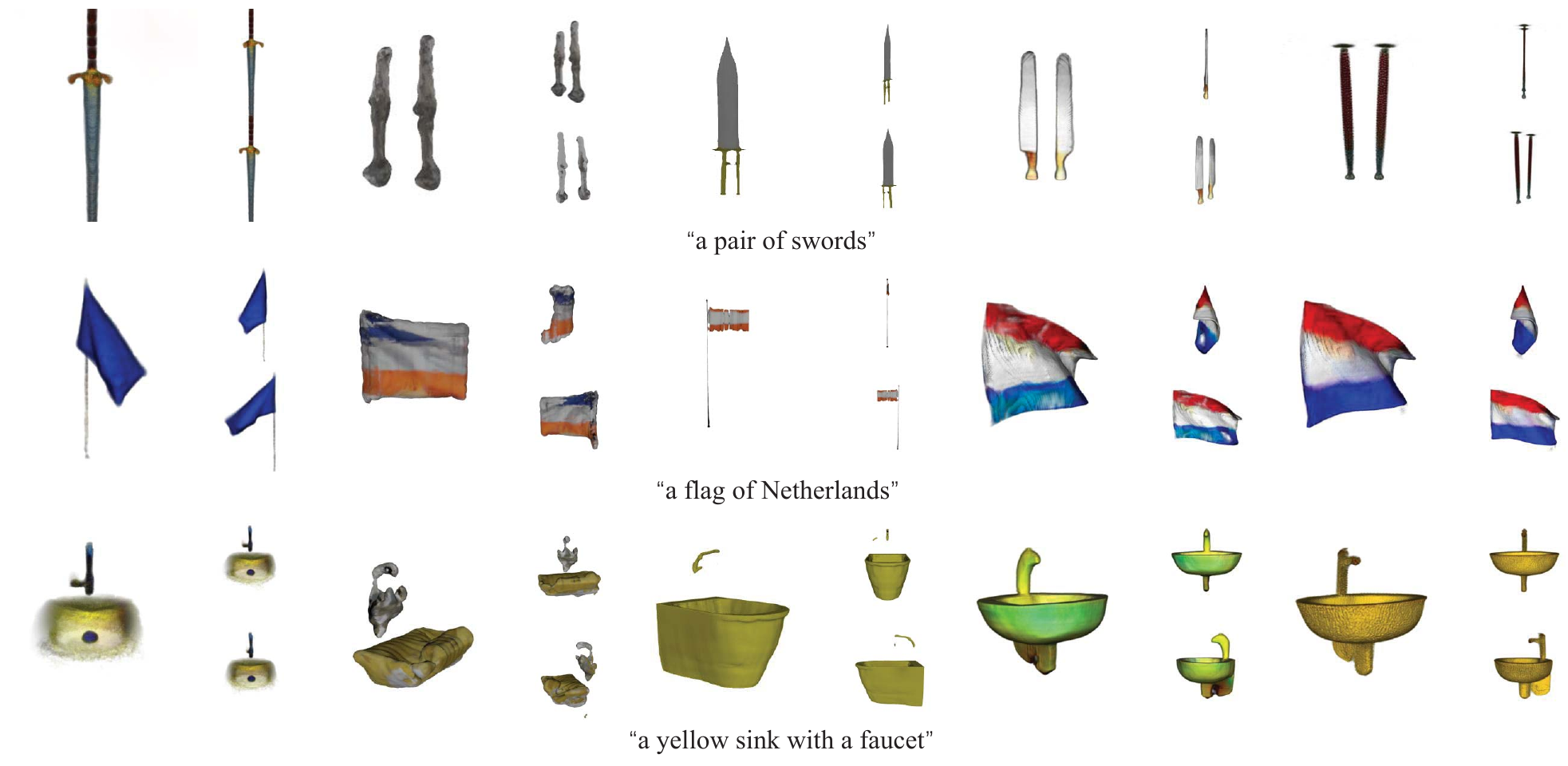}
   \includegraphics[width=1.0\linewidth,page=9]{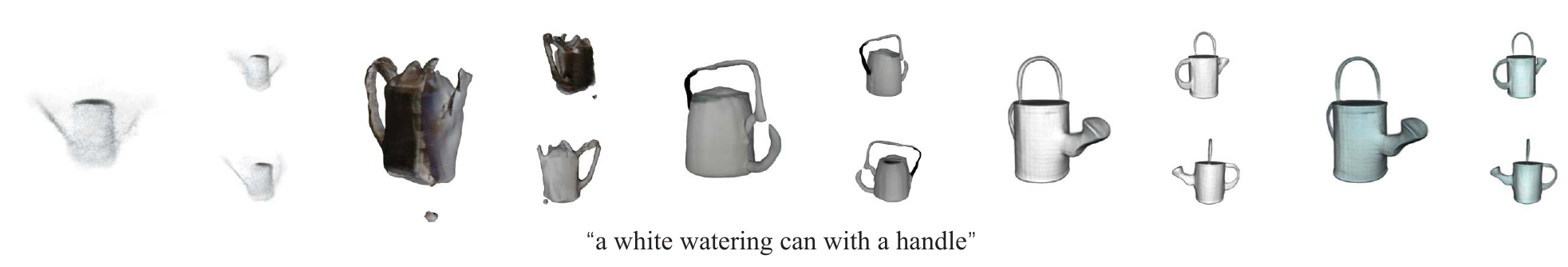}
   \caption{Comparison with state-of-the-art text-to-3D methods.}
   \label{fig:supp_comparison}
\end{figure*}

We present more results generated with our method in Figure~\ref{fig:supp_results_1} and Figure~\ref{fig:supp_results_2}. We emphasize the diversity in Figure~\ref{fig:supp_diversity} and the flexibility in Figure~\ref{fig:supp_flexibility} of our method. Also, we provide more comparisons with state-of-the-art text-to-3D approaches in Figure~\ref{fig:supp_comparison}.

\end{document}